\documentclass{article}
\usepackage[final]{neurips_2019}

\usepackage[utf8]{inputenc}
\usepackage[T1]{fontenc}   
\usepackage{hyperref}      
\usepackage{url}           
\usepackage{booktabs}      
\usepackage{amsfonts}      
\usepackage{nicefrac}      
\usepackage{microtype}     
\usepackage{graphicx}
\usepackage{caption}
\usepackage{subcaption}
\usepackage{amsmath}
\usepackage{amsthm}
\usepackage[ruled,boxed]{algorithm2e} 
\usepackage{algpseudocode}
\usepackage{natbib}
\usepackage{enumitem}

\usepackage[utf8]{inputenc}
\usepackage[english]{babel}
 
\newtheorem{theorem}{Theorem}
\newtheorem{definition}{Definition}

\newtheorem{corollary}{Corollary}

\setcitestyle{numbers}

\usepackage{xcolor}
\definecolor{green1}{RGB}{0,153,0}
\definecolor{red1}{RGB}{204,0,0}
\definecolor{blue1}{RGB}{0,0,153}
\usepackage{color}

\newcommand{\monetzero}{$\text{GloVe}_{meta}$}

\newcommand{\monetg}{$\text{MONET}_G$}
\newcommand{\leakage}{\mathcal{ML}}

\title{MONET: Debiasing Graph Embeddings via the Metadata-Orthogonal Training Unit}

\author{John Palowitch\\
Google Research\\
San Francisco, CA 94105\\
\texttt{palowitch@google.com}\\
\And
Bryan Perozzi\\
Google Research\\
New York, NY 10011\\
\texttt{bperozzi@acm.org} \\
}

\begin{document}

\maketitle

\begin{abstract}
In many real world graphs, the formation of edges can be influenced by certain sensitive features of the nodes (e.g.\ their gender, community, or reputation).
In this paper we argue that when such influences exist, any downstream Graph Neural Network (GNN) will be implicitly biased by these structural correlations.
To allow control over this phenomenon, we introduce the Metadata-Orthogonal Node Embedding Training (MONET) unit, a general neural network architecture component for performing training-time linear debiasing of graph embeddings.
MONET operates by ensuring that the node embeddings are trained on a hyperplane orthogonal to that of the node features (metadata).
Unlike debiasing approaches in similar domains, our method offers exact guarantees about the correlation between the resulting embeddings and any sensitive metadata.
We illustrate the effectiveness of MONET though our experiments on a variety of real world graphs against challenging baselines (e.g.\ adversarial debiasing), showing superior performance in tasks such as preventing the leakage of political party affiliation in a blog network, and preventing the gaming of embedding-based recommendation systems.
\end{abstract}

\section{Introduction}

\begin{figure}[!htb]
\centering
\includegraphics[scale=0.25]{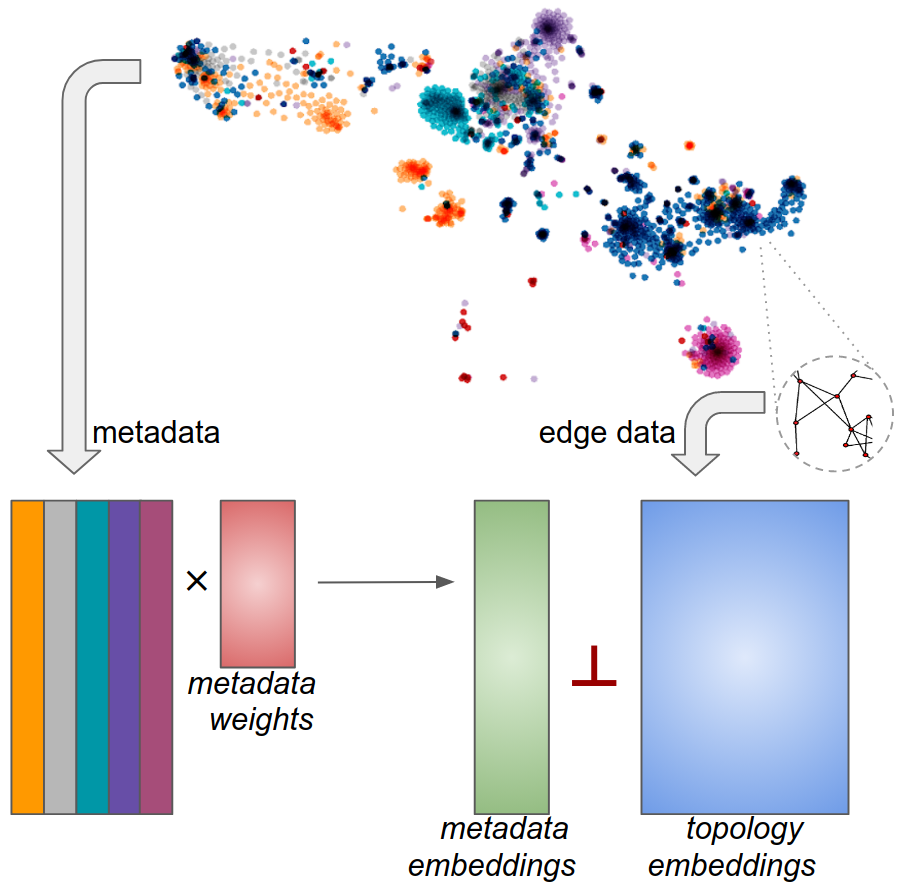}
\caption{Illustration of the Metadata-Orthogonal Node Embedding Training (MONET) model. Points are nodes from a toy graph, represented in 2-D, colored by known metadata. There is clear correlation and clustering by the metadata. MONET encodes the metadata signal, then debiases standard topology embeddings from the metadata embeddings. As described in Section \ref{sec:monet}, both representations are trained concurrently and orthogonally.}
\label{fig:monet-schematic}
\end{figure}

Graph embeddings -- continuous, low-dimensional vector representations of nodes -- have been eminently useful in network visualization, node classification, link prediction, and many other graph learning tasks \citep{cui2018survey}.
While graph embeddings can be estimated directly by unsupervised algorithms using the graph's structure  \citep[e.g.][]{perozzi2014deepwalk, tang2015line, grover2016node2vec,qiu2018network}, 
there is often additional  (non-relational) information available for each node in the graph.
This information, frequently referred to as node \emph{attributes} or node \emph{metadata}, can contain information that is useful for prediction tasks including  demographic, geo-spatial, and/or textual features.

The interplay between a node's metadata and edges is a rich and active area of research.
Interestingly, in a number of cases, this metadata can be measurably related to a graph's structure \citep{peel2017ground}, and in some instances there may be a causal relationship (the node's attributes influence the formation of edges).
As such, metadata can enhance graph learning models \citep{yang2015network, newman2016structure}, and conversely, graphs can be used as regularizers in supervised and semi-supervised models of node features \citep{yang2016revisiting, defferrard2016convolutional}. Furthermore, metadata are commonly used as evaluation data for graph embeddings \citep{chen2018tutorial}. For example, node embeddings trained on a Flickr user graph were shown to predict user-specified Flickr ``interests" \citep{perozzi2014deepwalk}. This is presumably because users (as nodes) in the Flickr graph tend to follow users with similar interests, which illustrates a potential causal connection between node topology and node metadata.

However, despite the usefulness and prevalence of metadata in graph learning, there are instances where it is desirable to design a system to \emph{avoid} the effects of a particular kind of \emph{sensitive} data.
For instance, the designers of a recommendation system may want to make recommendations independent of a user's demographic information or location.

At first glance, this may seem like an artificial dilemma -- surely one could just avoid the problem by not adding such sensitive attributes to the model.
However, such an approach (ignoring a sensitive attribute) does not control for any existing correlations that may exist between the sensitive metadata and the edges of a node.
In other words, if the edges of the graph are correlated with sensitive metadata, then any algorithm which does not explicitly model and remove this correlation will be biased as a result of it. 

To this end, we propose two simple desiderata for handling sensitive metadata in GNNs:
\setlist{nolistsep}
\begin{itemize}[noitemsep]
    \item[\textbf{D1}.] The influence of metadata on the graph topology is modeled in a partitioned subset of embedding space, providing interpretability to the overall graph representation, and the option to remove metadata dimensions.
    \item[\textbf{D2}.] Non-metadata or ``topology" embeddings are debiased from metadata embeddings with a \emph{provable guarantee} on the level of remaining bias.
\end{itemize}
Existing embedding methods for graphs with metadata \citep[e.g.][]{zhu2007combining,yang2015network,newman2016structure,zhang2016homophily,li2017ppne,guo2018enhancing} do not consider D2, and most do not not fully address D1 either. In the graph CNN literature \citep[e.g.][]{kipf2016semi,defferrard2016convolutional,abuelhaija2019mixhop}, metadata are treated as features in predictive models, rather than attributes to be controlled. The work most relevant to the above desiderata is \cite{bose2019compositional}, which introduces adversarial debiasing of metadata to graph-based recommender systems. However, that approach does not carry strong theoretical guarantees about the amount of bias removed from the graph representation, and therefore does not satisfy D2.

In this work we propose a novel technique for Graph Neural Networks (GNNs) that satisfies both of these desiderata. 
Our method, the Metadata-Orthogonal Node Embedding Training (MONET) unit, operates by ensuring that metadata embeddings are trained on a hyperplane orthgonal to that of the topology embeddings (as conceptually illustrated in Figure \ref{fig:monet-schematic}). 
Specifically, our contributions are the following:
\begin{enumerate}[noitemsep]
    \item The Metadata-Orthogonal Node Embedding Training (MONET) unit, a novel GNN algorithm which jointly embeds graph topology and graph metadata while enforcing linear decorrelation between the two embedding spaces.
    \item Analysis which proves that addressing desiderata D1 alone -- partitioning a metadata embedding space -- still produces a biased topology embedding space, and that the MONET unit corrects this.
    \item Experimental results on real world graphs which show that MONET can successfully debias topology embeddings while relegating metadata information to separate dimensions -- achieving superior results to state-of-the-art adversarial and NLP-based debiasing methods.
\end{enumerate}

\section{Preliminaries}
Early graph embedding methods involved dimensionality reduction techniques like multidimensional scaling and singular value decomposition \cite{chen2018tutorial}.
In this paper we use graph neural networks trained on random walks, similar to DeepWalk \cite{perozzi2014deepwalk}. DeepWalk and many subsequent methods first generate a sequence of random walks from the graph, to create a ``corpus" of node ``sentences" which are then modeled via word embedding techniques (e.g. word2vec \cite{mikolov2013distributed} or GloVe \cite{pennington2014glove}) to learn low dimensional representations that preserve the observed co-occurrence similarity.

Let $W$ be a $d$-dimensional graph \emph{embedding} matrix, $W\in \mathbb{R}^{n\times d}$, which aims to preserve the low-dimensional structure of a graph ($d<<n$). Rows of $W$ correspond to nodes, and node pairs $i,j$ with large dot-products $W_i^TW_j$ should be structurally or topologically close in the graph. As a concrete example, in this paper we consider the debiasing of a recently proposed graph embedding using the GloVe model \cite{brochier2019global}. Its training objective is:
\begin{equation}
    \label{eq:glove}
    L_{\text{GloVe}} = \sum_{i,j\leq n}f_\alpha(C_{ij})(a_i + b_j + U_i^TV_j -\log(C_{ij}))^2.
\end{equation}
Above, $U,V\in \mathbb{R}^{n\times d}$ are the ``center" and ``context" embeddings, $a, b\in \mathbb{R}^{n\times 1}$ are the biases, $C$ is the walk-distance-weighted context co-occurrences, and $f_\alpha$ is the loss smoothing function \cite{pennington2014glove}. We use the GloVe model in the next section, and throughout the paper, to illustrate the incorporation of metadata embeddings and the MONET unit. However, these innovations are generally deployable modules. To illustrate this, we also describe a MONET unit for DeepWalk \citep{perozzi2014deepwalk}, a popular graph embedding algorithm.

\textbf{Notation.} In this paper, given a matrix $A\in \mathbb{R}^{n\times d}$ and an index $i\in{1,\ldots, n}$, $A_i$ denotes the $d\times 1$ $i$-th row vector of $A$. Column indices will not be used. $\mathbf{0}_{n\times d}$ denotes the $n\times d$ zero matrix, and $||\cdot||_F$ denotes the Frobenius norm.

\section{Metadata Embeddings and Orthogonal Training}\label{sec:method}

In this section we present the components of MONET. First, in Section \ref{sec:naivemodel}, we extend traditional graph representation to include metadata embeddings, achieving desiderata D1. Next, in Section \ref{sec:leakage}, we prove that a model with just D1 will still leak metadata information to the topology embeddings. Then, in Section \ref{sec:monet} we present MONET's key algorithm for training metadata and topology embeddings orthogonally, achieving desiderata D2. Finally, we conclude with some analysis of MONET in Section 3.4.

\subsection{Jointly Modeling Metadata \& Topology}
\label{sec:naivemodel}
\begin{figure}[!htb]
\centering
\includegraphics[scale=0.4]{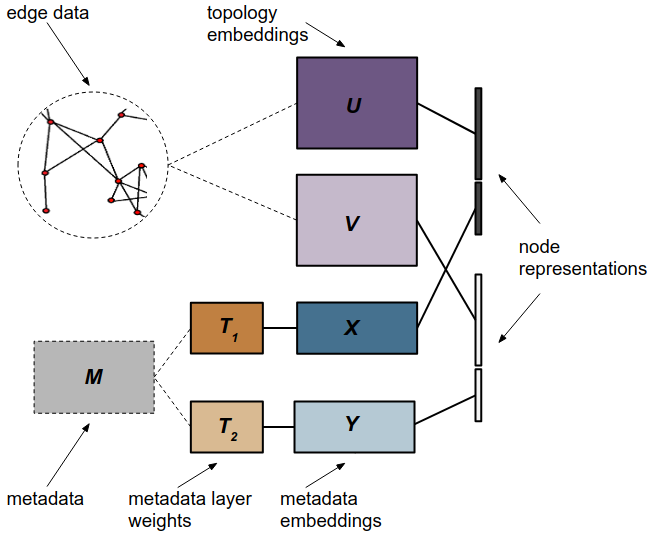}
\caption{ Illustration of \monetzero. $U$ and $V$ are topology embedding layers. A single-layer feed-forward neural network creates metadata embeddings layer $X$ and $Y$. The concatenations $[U, X]$ and $[V, Y]$ create the output node representations.
\label{fig:glove-meta-architecture}
}
\end{figure}

A natural first approach to controlling metadata effects is to introduce a partition of the embedding space that models them explicitly, adding interpretability and separability to the overall representation. 
We denote the node metadata matrix as $M\in R^{n\times m}$, where the metadata for node $i$ is contained in the row vector $M_i$.
To achieve D1, we feed $M$ through a single-layer neural network with weights $T$.
Figure \ref{fig:glove-meta-architecture} shows a realization of this idea, using the GloVe embedding model, which we denote \monetzero.  Here, two weight matrices $T_1, T_2$ are needed to embed the metadata for both center and context nodes.
This network yields metadata \emph{embeddings} $X$, $Y$, which correspond to the standard topology embeddings $U$, $V$ (respectively), and can be concatenated to form a combined metadata and topology vector. 
The resulting joint metadata and topology loss is:
\begin{equation}\label{eq:lmeta}
    L_{\text{GloVe\textsubscript{meta}}} = \sum_{i,j\leq n}f_\alpha(C_{ij})(a_i + b_j + U_i^TV_j + X_i^TY_j-\log(C_{ij}))^2. 
\end{equation}

While in this paper we demonstrate metadata embeddings using GloVe, we note that they can be incorporated in any GNN which utilizes a node-wise graph representation. 
For instance, the well-known DeepWalk \citep{perozzi2014deepwalk} loss, which is based on word2vec \citep{mikolov2013distributed}, would incorporate metadata embeddings as follows:
\begin{equation*}
    -\sum_{i,j\in\mathcal{W}}\log(U_i^TV_j + X_i^TY_j) - \sum_{k\in K_i}\log(-U_i^TV_k - X_i^TY_k).
\end{equation*}
Above, $\mathcal{W}$ is the set of context pairs from random walks, and $K_i$ is a set of negative samples associated with node $i$.

This new modeling approach, which augments standard graph representations with a metadata partition, provides users of graph embeddings with the option to include or exclude metadata-learned dimensions. Furthermore, suppose that the metadata (e.g.\ demographic information) are indeed associated with the formation of links in the graph. 
In this case, ostensibly, the dedicated metadata embedding space could relieve the topology dimensions of the responsibility to encode this relationship, thereby reducing metadata bias in those dimensions.
In our empirical study (Section 4) we show that to some extent, this does actually occur. However, we find -- empirically and theoretically -- that this na\"{\i}ve approach does not guarantee that the topology embeddings are completely decorrelated from the metadata embeddings. In fact, suprisingly, we find that most of the metadata bias sufferred by standard baselines remain in the topology dimensions. 
This is a phenomenon we call \emph{metadata leakage}, which we formalize and prove in the next section.

\subsection{Metadata Leakage in Graph Neural Networks}
\label{sec:leakage}
Here, we formally define \emph{metadata leakage} for general topology and metadata embeddings, and show how it can occur even in embedding models with partitioned metadata embeddings. This motivates the need for both D1 and D2 in a complete approach to controlling metadata effects in GNNs. All proofs appear in the supplement.

\begin{definition} The \textbf{metadata leakage} of metadata embeddings $Z\in\mathbb{R}^{n\times d_Z}$ into topology embeddings $W\in\mathbb{R}^{n\times d}$ is defined $\leakage(Z, W) := ||Z^TW||_F^2$. We say that there is no metadata leakage if and only if $\leakage(Z,W)=0$.
\end{definition}

Without a more nuanced approach, metadata leakage can occur even in embedding models with a metadata partition, like those discussed in Section \ref{sec:naivemodel}. To demonstrate this we consider a reduced metadata-laden GloVe loss with unique topology and metadata representations $W$ and  $Z=MT$:
\begin{align}\label{eq:lmeta-red}
L^{\ast}_{\text{GloVe\textsubscript{meta}}} = \sum_{i,j\leq n}f_\alpha(C_{ij})(a_i + a_j + W_i^TW_j + Z_i^TZ_j -\log(C_{ij}))^2
\end{align}
We now show that under a random update of GloVe under Eq.\ \eqref{eq:lmeta-red}, the expected metadata leakage is non-zero. Specifically, let $(i, j)$ be a node pair from $C$, and define $\delta_W(i,j)$ as the incurred Stochastic Gradient Descent update $W'\leftarrow W + \delta_W(i,j)$. Suppose there is a ``ground-truth" metadata transformation $B\in \mathbb{R}^{m\times d_B}$, and define ground-truth metadata embeddings $\tilde{Z}:= MB$, which represent the ``true" dimensions of the metadata effect on the co-occurrences $C$. Define $\Sigma_B:=BB^T$ and $\Sigma_T:=TT^T$. With expectations taken with respect to the sampling of a pair $(i, j)$ for Stochastic Gradient Descent, define $\mu_W:=\mathbb{E}[W_i]$ and $\Sigma_W := \mathbb{E}[W_iW_i^T]$. Define $\mu_M$, $\Sigma_M$ similarly. Then our main Theorem is as follows:
\begin{theorem}
Assume $\Sigma_W = \sigma_WI_{d}$ for $\sigma_W>0$, $\mu_W = \textbf{0}_{d\times 1}$, and $\mu_M = \textbf{0}_{m\times 1}$. Suppose for some fixed $\theta \in \mathbb{R}$ we have $\log(C_{ij}) = \theta + \tilde{Z}_i^T\tilde{Z}_j$. Let $(i,j)$ be a randomly sampled co-occurrence pair and $W'$ the incurred update. Then if $\mathbb{E}[M_iW_i^T] = \beta\in\mathbb{R}^{m\times d}$, we have
\begin{align}
    \mathbb{E}[\leakage(Z, &W')] \geq 2||T^T\left[\Sigma_M(\Sigma_B - \Sigma_T) + (n - \sigma_W)I_{m}\right]\beta||_F^2.
\end{align}
\end{theorem}

Importantly, $\Sigma_T$ and $\sigma_W$ are neural network hyperparameters, so we give a useful Corollary:
\begin{corollary}
Under the assumptions of Theorem 1, $\mathbb{E}[\leakage(Z, W)] = \Omega( n||T^T\beta||_F^2 )$ as $n\rightarrow\infty$.
\end{corollary}
Note that under reasonable GNN initialization schemes, $T$ and $\beta$ are random perturbations. Thus, Corollary 1 implies the surprising result that incorporating a metadata embedding partition is not sufficient to prevent metadata leakage in practical settings.

\subsection{MONET: Metadata-Orthogonal Node Embedding Training}
\label{sec:monet}

Here, we introduce the  Metadata-Orthogonal Node Embedding Training (MONET) unit for training joint topology-metadata graph representations $[W, Z]$ without metadata leakage. MONET explicitly prevents the correlation between topology and metadata, by using the Singular Value Decomposition (SVD) of $Z$ to \emph{orthogonalize} updates to $W$ during training.

\begin{figure}[!htb]
\centering
\includegraphics[scale=0.4]{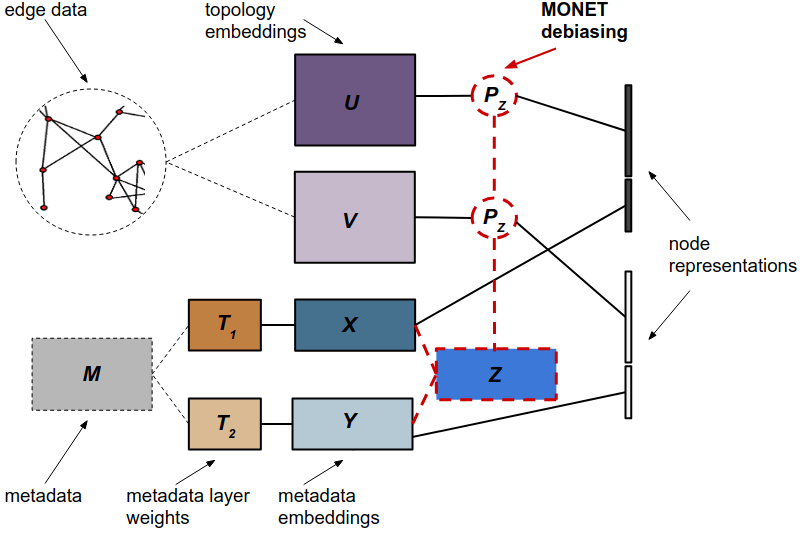}
\caption{\label{fig:monetg} The MONET unit adds a feed-forward transformation of the metadata, resulting in metadata embeddings $X$ and $Y$. $Z = X+Y$ gives the combined metadata representation, used to debias $U$ and $V$ via $P_Z$.}
\end{figure}

\textbf{MONET.} The MONET unit is a two-step algorithm applied to the training of a topology embedding in a neural network, and is detailed in Algorithm \ref{alg:monet}.
The input to a MONET unit is a metadata embedding $Z\in\mathbb{R}^{n\times d_z}$ and a target topology embedding $W\in\mathbb{R}^{n\times d}$ for debiasing.
Then, let $Q_Z$ be the left-singular vectors of $Z$, and define the projection $P_Z := I_{n\times n} - Q_ZQ_Z^T$. In the forward pass procedure, debiased topology weights are obtained by using the projection $W^\perp = P_ZW$. Similarly, $W^\perp$ is used in place of $W$ in subsequent GNN layers. 
In the backward pass, MONET also debiases the backpropagation update to the topology embedding, $\delta_W$, using $\delta_W^\perp=P_Z\delta_W$.

\begin{algorithm}[t]
\caption{MONET Unit Training Step}\label{alg:monet}
\begin{algorithmic}[1]
\State Input: topology embedding $W$, metadata embedding $Z$
\Procedure{Forward Pass debiasing}{$W$, $Z$}
\State Compute $Z$ left-singular vectors $Q_Z$ and projection: \newline \indent $P_Z\gets I_{n\times n} - Q_ZQ_Z^T$
\State Compute orthogonal topology embedding: \newline \indent $W^{\perp} \gets P_ZW$
\State \Return debiased graph representation $[W^\perp, Z]$
\EndProcedure
\Procedure{Backward Pass debiasing}{$\delta_W$}
\State Compute orthogonal topology embedding update: \newline \indent $\delta^\perp_W \gets P_Z\delta_W$
\State Apply update: \newline \indent $W^\perp\gets W^\perp + \delta^\perp_W$
\State \Return debiased topology embedding $W^\perp$
\EndProcedure
\end{algorithmic}
\end{algorithm}

Straightforward properties of the SVD show that MONET directly prevents metadata leakage:
\begin{theorem}
Using Algorithm \ref{alg:monet}, $\leakage(Z, W^\perp) = 0$ and $\leakage(Z, \delta^\perp) = 0$.
\end{theorem}
We note that in this work we have only considered \emph{linear} metadata leakage; provable guarantees for debiasing nonlinear topology/metadata associations is an area of future work.

\textbf{Implementation (\monetg).}
We demonstrate MONET in our experiments by applying Algorithm 1 to Eq. \eqref{eq:lmeta}, which we denote as \monetg.
We orthogonalize the input and output topology embeddings $U,V$ with the summed metadata embeddings $Z := X + Y$. By linearity, this implies $Z$-orthogonal training of the summed topology representation $W=U+V$. We note that working with the sums of center and context embeddings is the standard way to combine these matrices \citep{pennington2014glove}. Figure \ref{fig:monetg} gives a full illustration of \monetg.

\textbf{Relaxation}. A natural generalization of the MONET unit is to parameterize the level of orthogonalization incurred at each training step. Specifically, we introduce a parameter $\lambda\in[0.0, 1.0]$ which controls the extent to which topology embeddings are projected onto the metadata-orthogonal hyperplane:
\begin{equation}
    P^{(\lambda)}_Z := I_{n\times n} - \lambda Q_ZQ_Z^T
\end{equation}
$P^{(\lambda)}_Z$ takes the role of $P_Z$ in \ref{alg:monet}. Using MONET with $\lambda = 1.0$ removes all linear bias, whereas using $\lambda = 0.0$ prevents any debiasing. In Section 4.2, we explore how $\lambda$ affects the trade-off between linear debiasing and task accuracy.

\subsection{Analysis}
\label{sec:analysis}

We briefly remark about the algorithmic complexity of MONET, and the interpretation of its parameters.

\textbf{Algorithmic Complexity.} The bottleneck of MONET occurs in the SVD computation and orthogonalization. In our setting, the SVD is $O(nd_z^2)$ \cite{trefethen1997numerical}. The matrix $P_Z$ need not be computed to perform orthogonalization steps, as $P_ZW = W - Q_Z(Q_Z^TW)$, and the right-hand quantity is $O(ndd_z)$ to compute. Hence the general complexity of the MONET unit is $O(nd_z\max\{d, d_z\})$. In the experiments section we compare the wall clock time of MONET and baselines, showing only about a 23\% overall wall time increase from standard GloVe.

\textbf{Metadata Parameter Interpretation.} The $i,j$ terms in the sum of the loss for GloVe models with metadata (\monetzero\; and \monetg) involve the dot product $X_i^TY_j = M_i^TT_1T_2^TM_j$. That expansion suggests that the matrix $\Sigma_T := T_1T_2^T$ contains all pairwise metadata dimension relationships. In other words, $\Sigma_T$ gives the direction and magnitude of the raw metadata effect on log co-occurrence, and is therefore a way to measure the extent to which the model has captured metadata information. We will refer to this interpretation in the experiments that follow. An important experiment will show that applying the MONET algorithm increases the magnitude of $\Sigma_T$ entries.

\section{Experimental Analysis}

In this section we design and analyze experiments with the goal of answering the following questions:
\begin{itemize}
\item[Q1.] Can MONET remove linear bias from topology embeddings and downstream prediction tasks?
\item[Q2.] Can MONET help reduce bias in topology embeddings even when they are used in nonlinear models?
\item[Q3.] Can MONET provide a reasonable trade-off between debiasing and accuracy on a downstream retrieval task?
\item[Q4.] Does adding the MONET unit to a method incur a significant performance overhead?
\end{itemize}

In all experiment settings, our topology embedding of interest for evaluation is the sum $W:=U+V$ of the center-context topology embeddings $U$ and $V$. All GloVe-based models are trained with TensorFlow \citep{abadi2016tensorflow} using the AdaGrad optimizer \citep{duchi2011adaptive} with initial learning rate 0.05 and co-occurrence batch size 100. DeepWalk was trained using the gensim software \citep{rehurek_lrec}. Software for the MONET algorithm and code to reproduce the following experiments are available in the supplemental material\footnote{Software and analysis code also available at \url{http://urlremovedforreview}}.

\subsection{Experiment 1: Debiasing a Political Blogs Graph}
Here we seek to answer \textbf{Q1} and \textbf{Q2}, illustrating the effect of MONET debiasing by
removing the effect of political ideology from a blogger network \cite{adamic2005political}. The political blog network\footnote{Available within the Graph-Tool software \citep{peixoto_graph-tool_2014}} has has 1,107 nodes corresponding to blog websites, 19,034 hyperlink edges between the blogs (after converting the graph to be undirected), and two clearly defined, equally sized communities of liberal and conservative bloggers.

\textbf{Baseline Methods}. In this experiment we compare \monetg\ against the following methods: (1) a random topology embedding generated from a multivariate Normal distribution; (2) DeepWalk; (3) GloVe; (4) \monetzero; (5) an adversarial version of GloVe, in which a 2-layer MLP discriminator is trained to predict political affiliation  \cite{goodfellow2014generative,bose2019compositional}. All methods use 16-dimensional topology embeddings and (if appropriate) 2-dimensional metadata embeddings of blog affiliation. Full details on methods and training are given in the supplement.

\textbf{Design}. In this experiment we use two metrics to measure the bias of topology embeddings $W$:
\begin{itemize}[noitemsep]
    \item Accuracy of a blog affiliation classifier trained on $W$.
    \item Metadata Leakage $\leakage(W, Z)$.
\end{itemize}
Because we are studying debiasing potential, Accuracy scores close to 0.5 are desirable (due to equally-sized blog affiliation communities), and Metadata Leakage equal to 0.0 is optimal. For methods without metadata embeddings, $\leakage$ is computed with the original metadata $M$.

One repetition of the experiment proceeds as follows. Graph representations are trained with each method (some include metadata embedding partitions). For each $p\in\{10, 20, \ldots, 90\}$, a train set of $p\%$ of nodes is sampled uniformly. Classifiers of blog affiliation are trained using each methods' topology embeddings. We compute Accuracy on the test set and $\leakage$ on all nodes. We report mean and standard deviations across 10 independent repetitions.

\begin{figure}[!t]
\begin{subfigure}[t]{0.5\textwidth}
    \includegraphics[scale=0.17]{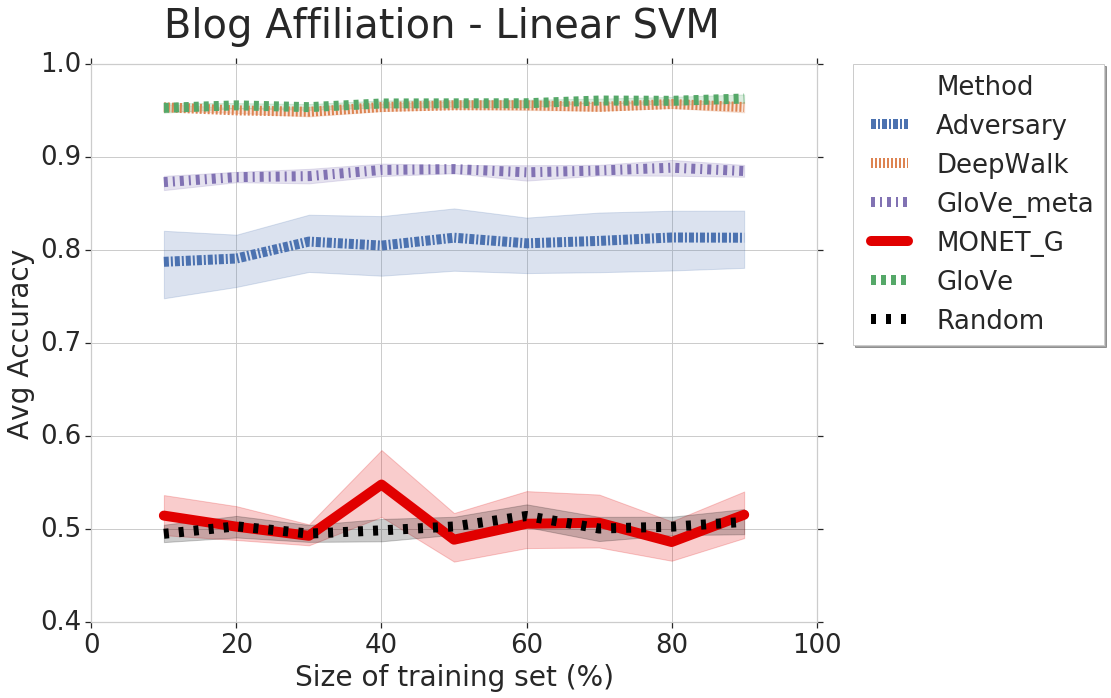}
    \caption{Sensitive attribute prediction using linear model}
    \label{fig:svm-results-linear}
\end{subfigure}
\begin{subfigure}[t]{0.5\textwidth}
    \includegraphics[scale=0.17]{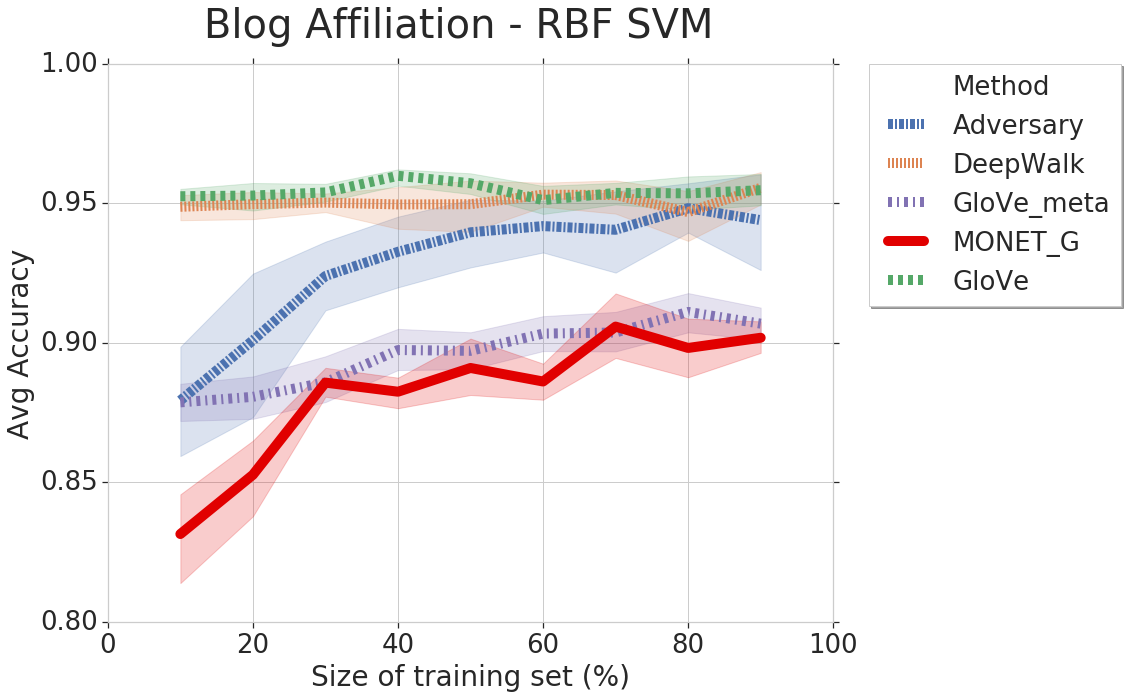}
    \caption{Sensitive attribute prediction using nonlinear model}
    \label{fig:svm-results-rbf}
\end{subfigure}
\caption{\label{fig:svm-results} Accuracy results from blog political affiliation classifiers.  Lower is better (embeddings more debiased). \monetg\ achieves this perfectly with a linear classifier, due to the MONET unit's novel orthogonalization technique. Interestingly, on this task, \monetg\ is dominant even under a nonlinear classifier.}
\end{figure}

\textbf{Results}. 
First we analyze the accuracy of a linear SVM at predicting political party affliation, shown in Figure \ref{fig:svm-results-linear}. 
The performance of \monetg\ under the linear classifier is indistinguishable from the random baseline, which shows that the MONET unit perfectly debiased the GloVe embeddings during training, answering question \textbf{Q1} in the affirmative.  We note that this result is a direct implication of Theorem 2.

Interestingly, although the naive \monetzero\ also has a metadata embedding partition, its performance under the linear classifier is nearly the same as baselines with no metadata (GloVe and DeepWalk). This demonstrates the downstream effect of Theorem 1, and shows the necessity of desiderata D2 in a rigorous approach to graph embedding debiasing. We note that MONET also outperforms the adversarial extension of GloVe, even though adversarial approaches have performed well in similar domains \citep{bose2019compositional}.

Second, we observe the results from training a non-linear classifier (the RBF SVM), as shown in Figure \ref{fig:svm-results-rbf}.
Here, we see that all embedding methods contain non-linear biases to exploit for the prediction of political affiliation. 
This includes \monetg\, because the MONET unit only performs linear debiasing. 
Surprisingly, here we find that \monetg\ still produces the \emph{least} biased representations, even beating an adversarial approach.
This answers question \textbf{Q2}, showing that MONET can reduce bias even when its representations are used as features inside of a non-linear model. 

\begin{table}[!t]
    \centering
    \begin{tabular}{c|c|c}
    \hline
    Method & $\leakage$ & Metadata Importance $\Sigma_T$\\\hline
    GloVe & 6503 $\pm$ 196 & N/A \\
    Random & 283 $\pm$ 22 & N/A\\
    Adversary & 4430 $\pm$ 305 & N/A\\
    DeepWalk & 2670 $\pm$ 104 & N/A\\
    \monetzero\ & 2103 $\pm$ 183 & $\begin{pmatrix}.110 \pm .01 & -.106 \pm .01\\ -.110 \pm .01 & .106 \pm .01\end{pmatrix}$\\
    \monetg\ & $\mathbf{.021 \pm .003}$ & $\begin{pmatrix}.187 \pm .01 & -.182 \pm .01\\ -.187 \pm .01 & .182 \pm .01\end{pmatrix}$\\\hline
    \end{tabular}
    \caption{\label{tab:leakage-table} When trained to debias political affiliation, MONET removes metadata leakage to precision error. Metadata Importance values - computed from layers of \monetzero\ and \monetg\ - show that the MONET unit allows stronger metadata learning.
    }
\end{table}

\textbf{Empirical Analysis}. Here we provide in-depth model understanding of MONET and, by comparison, \monetzero\ and GloVe. 
The metadata leakage ($\leakage$), as defined in Section \ref{sec:leakage}, is shown for each embedding set in Table \ref{tab:leakage-table}. We see that embedding models without metadata control have high leakage. 
As predicted by Corollary 1, \monetzero's metadata leakage also remains large -- whereas, due to Theorem 2, \monetg's $\leakage$ is at machine precision. This again shows that a metadata embedding partition is not sufficient to isolate the metadata effect. 

\begin{figure}[t!]
\centering
\mbox{
  \includegraphics[scale=0.20]{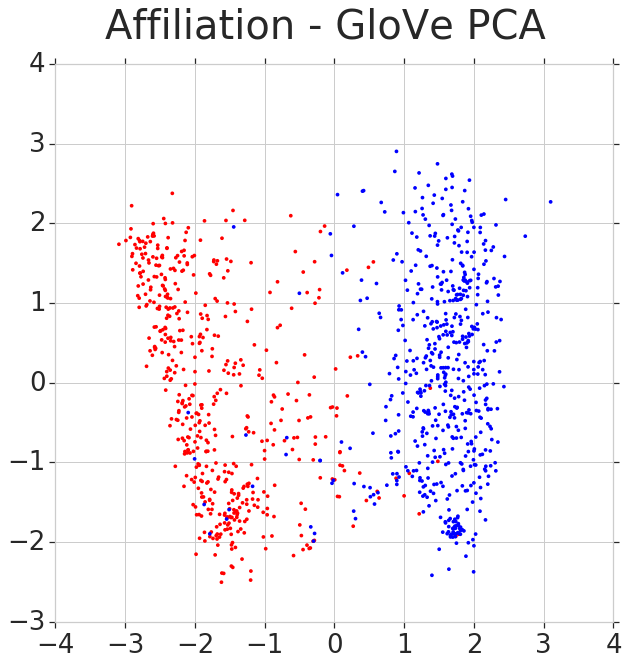}
  \includegraphics[scale=0.20]{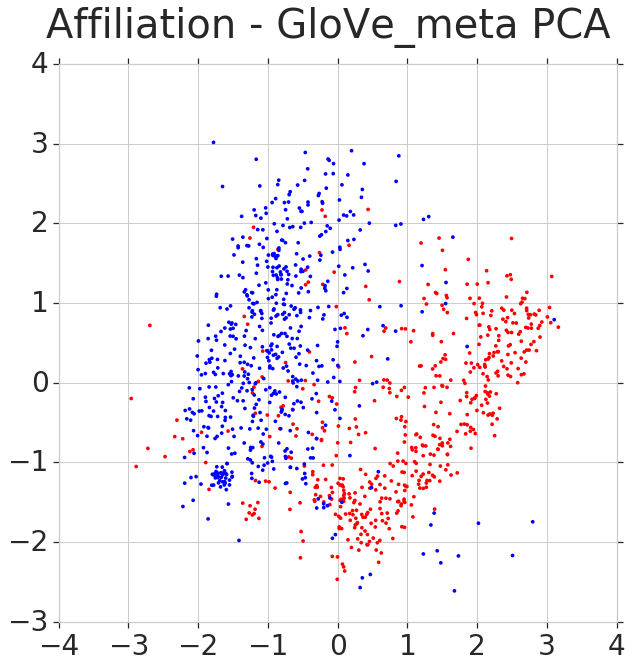}
  \includegraphics[scale=0.20]{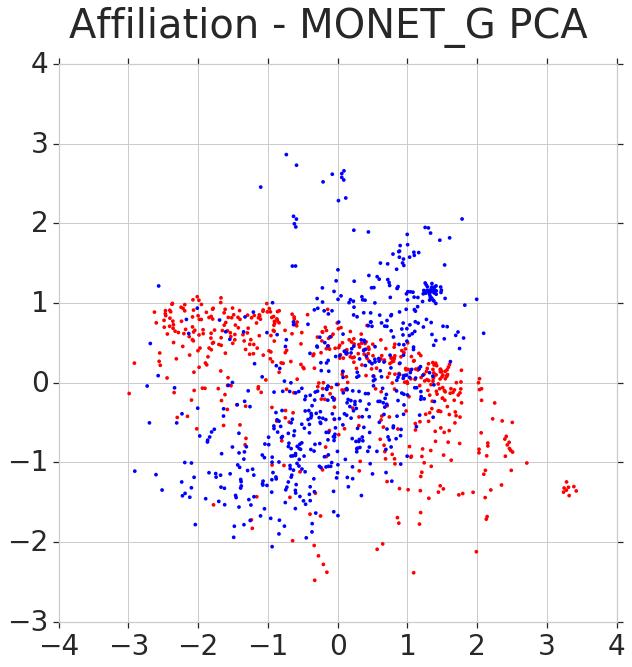}
}
\caption{PCA of political blog graph embeddings. (a): affiliation separation clearly visible on standard GloVe embeddings. (b): affiliation separation reduces when \monetzero\ captures some metadata information. (c): affiliation separation disappears with \monetg\ orthogonalized training.}
\label{fig:polblogs-affiliation}
\end{figure}

The need for \monetg\ over the na\"{\i}ve \monetzero\ can be observed in two other ways. First, recall from Section \ref{sec:analysis} that the ``Metadata Importance" matrix $\Sigma_T$ encodes pairwise relationships between metadata dimensions in embedding space. There is a noticeable increase in $\Sigma_T$ magnitude when MONET is used, implying that \monetzero\ metadata embeddings are not capturing all possible metadata information. Second, Figure \ref{fig:polblogs-affiliation} shows the 2-D PCA of each models' 16-dimensional topology embeddings, colored by political affiliation. Clear separation by affiliation is visible with the GloVe model's PCA. The PCA of \monetzero\ also shows separation, but less so. This shows that having a metadata partition in embedding space -- desiderata D1 -- does some work toward debiasing the topology dimensions. However, the orthogonalizing MONET unit does the bulk of the work, as shown by the overlap of the affiliation clusters in the \monetg\ PCA.

\subsection{Experiment 2: Debiasing a Shilling Attack}

In this experiment we investigate \textbf{Q3} and \textbf{Q4} in the context of using debiasing methods to defend against a \emph{shilling attack} \citep{Chirita:2005:PSA:1097047.1097061} on graph-embedding based recommender systems \citep{ying2018graph}. In a shilling attack, a number of users act together to artificially increase the likelihood that a particular influenced item will be recommended for a particular target item.

\textbf{Data.} In a single repetition of this experiment, we inject an artificial shilling attack into the MovieLens 100k dataset\footnote{\url{http://files.grouplens.org/datasets/movielens/ml-100k/}}. The raw data is represented as a bipartite graph with 943 users, 1682 movies (``items"), and a total of 100,000 ratings (edges). Each user has rated at least 20 items. At random, we sample 10 items into an influence set $S_I$, and a target item $i_t$ to be attacked. We take a random sample of 5\% of the existing users to be the set of attackers, $\mathcal{S}_A$. We then create a new graph, $G_\text{attacked}$ which in addition to all the existing ratings, contains new ratings from each attacker $\in \mathcal{S}_A$ to each item $\in \mathcal{S}_I$ as well as the target video.

\textbf{Methods}. We include all methods from the political blogs experiment in Section 4.1, with the exception of the adversarial baseline, as it was not implemented for continuous metadata. Instead, we applied a generalized correlation removal framework developed for removing word embedding bias \citep{schmidt2015rejecting,bolukbasi2016man}. Specifically, we compute an ``attack" direction (analogous to ``gender" direction in language modeling literature) as follows. Let $W$ be the GloVe topology embedding, and let $M$ be the attacker metadata described above. Note that $M_i$ is simply the scalar number of known attackers on item $i$. Then we compute the weighted attack direction as:
\begin{equation}
    \vec{a} := \sum_{i: M_i > 0}  M_i\cdot W_i - \sum_{i: M_i = 0}W_i
\end{equation}
Following \cite{schmidt2015rejecting}, we compute the ``NLP" baseline debiased graph embeddings as
\begin{equation}
    W_{\text{NLP}} := W - (W\vec{a}/||\vec{a}||)\vec{a}^T
\end{equation}
We also run relaxed \monetg\ (see Section 3.3) for various values of $\lambda$, to investigate the trade-off between accuracy and debiasing. 

As we wish to study item recommendation, in the random walks, we simply remove user nodes each time they are visited (so the walks contain only pairwise co-occurrence information over items). All methods compute 128 dimensional topology embeddings for the items. As metadata, we allow MONET models to know the per-item attacker rating count for each attacked item. However, to better demonstrate real-world performance, we only allow 50\% (randomly sampled) attackers from the original 5\% sample to be ``known" when constructing these metadata. Additional training details are given in the supplemental.

\textbf{Design}. In this experiment we compute two metrics to investigate the bias-accuracy trade-off:
\begin{itemize}
    \item The number of influence items in $S_I$ in top-20 embedding-nearest-neighbor list of $i_t$. This is a measure of \emph{downstream} bias on a retrieval task.
    \item The mean reciprocal rank for item retrieval. Let $NN(i)$ be the nearest neighbor of $i$ by random-walk co-occurrence count. Let $\text{rank}_i(j)$ be the integer rank of item $j$ against $i$'s complete set of embedding cosine distances. Then
    \begin{equation}
        \text{MRR} = n^{-1}\sum_i \text{rank}_i(NN(i))^{-1}
    \end{equation}
\end{itemize}

\begin{figure}[!t]
\centering
\includegraphics[scale=0.25]{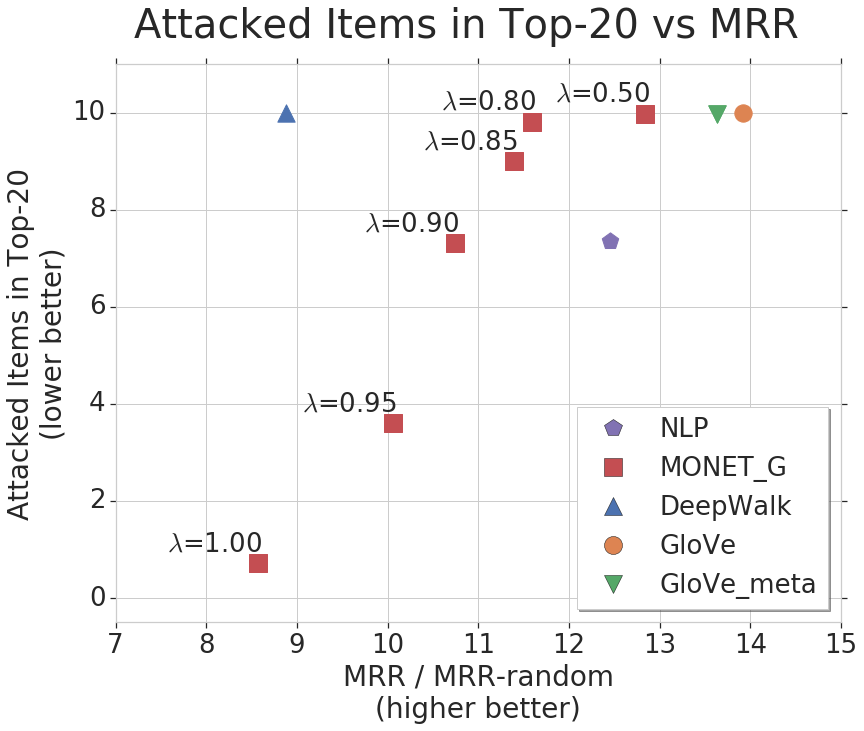}
\caption{\label{fig:mrr} Shilling experiment accuracy-bias trade-off: the $x$-axis is the lift-above-random of the mean reciprocal rank (MRR) of embedding-based item retrieval against item random-walk co-occurrence ordering. The $y$-axis is the number of attacked items in the top-20 embedding-distance nearest-neighbors of the target item $t_i$. All points are averages over ten runs of the shilling experiment. Only \monetg\ completely prevents attack bias, and maintains comparable performance to DeepWalk and greater than 8x better than a random baseline. Confidence intervals along both axes are provided in a supplemental table.}
\end{figure}

\textbf{Results.} The results of the shilling experiment are shown in Figure \ref{fig:mrr}, which compares the number of top-20 attacked items (a measure of bias) against MRR lift over a random baseline (a measure of accuracy).

First, we analyze how effectively each method debiased the embeddings.  Here we see that the topology embeddings from \monetg($\lambda=1.0$) prevent the most attacker bias by a large margin, letting less than $1$ influenced item into the top-20 neighbors of $t_i$.
We note that this behavior occurs even though the majority of observed co-occurrences for the algorithm had nothing to do with the attack in question, and only half of the true attackers were known. All other baselines (including those that explicitly model the attacker metadata) left at least around half of the attacked items in the top-20 list.  

Next, we consider the ranking accuracy of each method.  
Here we observe that \monetg\ outperforms the random baseline by at least 8.5x, and is comparable to one baseline (DeepWalk). 
Regarding the NLP debiasing method, we see that it outperformed \monetg\ in terms of ranking accuracy, but allowed a surprising number of attacked items ($\sim$7.5) in top-20 lists. We note that this method can not reduce bias further, and does offer  any guarantees of performance.
Finally, we observe that relaxing MONET's orthogonality constraint ($\lambda$, the level of orthogonal projection), decreases the effective debiasing, but improves embedding performance (as might be expected). Note that $\lambda = 0.0$ (not shown), recovers the \monetzero\ solution. This reveals a trade-off between embedding debiasing and prediction efficacy which has also been observed in other contexts \cite{bose2019compositional}. 
This confirms \textbf{Q3} -- that MONET allows for a controllable trade-off between debiasing and accuracy.

In addition to the ranking metrics, we also compute the Pearson correlation of each embedding w.r.t.\ the GloVe topology embeddings, shown in Table \ref{tab:timing}.
We find that \monetg\ achieves comparable correlation to \monetzero\ and NLP, showing that the SVD operation does not harshly corrupt the overall GloVe embedding space. 
We also report wall times for each method, showing that the metadata embedding and SVD operations of \monetg\ add negligibly to the runtime over GloVe.
Together, these metrics answer \textbf{Q4} in the negative.

\begin{table}[!t]
\centering
\begin{tabular}{c|c|c}
\hline
Method & \centering $W_{\text{GloVe}}$ Distances\ $r$ & Wall Time (sec) \\\hline
DeepWalk & 0.347 $\pm$ 0.003 & 45 $\pm$ 4 \\
GloVe & 1.000 $\pm$ 0.000 & 357 $\pm$ 54 \\
\monetzero & 0.790 $\pm$ 0.002 & 359 $\pm$ 49 \\
\monetg & 0.758 $\pm$ 0.004 & 442 $\pm$ 58 \\
Random & 0.031 $\pm$ 0.001 & N/A \\
NLP & 0.770 $\pm$ 0.008 & N/A \\\hline
\end{tabular}
\caption{\label{tab:timing} Shilling experiment performance metrics: ``$W_{\text{GloVe}}$ Distances\ $r$'' for a method is the Pearson correlation between (1) the cosine embedding distances produced by GloVe topology embeddings and (2) those same distances produced by the method. These figures show that \monetg\ does not overly corrupt the GloVe embedding signal, and does not add prohibitive computation time.}
\end{table}

\section{Conclusion}

This work introduced MONET, a novel GNN training technique which theoretically guarantees linear debiasing of graph embeddings from sensitive metadata. Our analyses illustrate the complexities of modeling metadata in GNNs, revealing that simply partitioning the embedding space to accommodate metadata does not achieve debiasing. In an empirical study, only MONET fully masked metadata signal from a linear classifier, and even outperformed adversarial techniques when used with a nonlinear model. A real-world item retrieval experiment showed that MONET successfully prevents shilling attack bias without severely impacting the performance of the underlying GNN.

There are two promising directions of future work in this new area. First, MONET only guarantees linear debiasing. Methods and exact guarantees for controlling nonlinear associations should be investigated. Toward this, the RBF SVM accuracy results on our political blogs experiment can serve as a benchmark. Second, we did not explore the performance of MONET in deeper or supervised GNNs, or its handling of high-dimensional metadata, each of which present interesting modeling challenges. For instance, in a deep supervised graph GNNs, it is not clear which hidden (embedding) layers would be worth debiasing, or whether associations between the metadata and prediction task are detrimental. While our paper makes a strong case for the use of MONET in current industry applications, answering these questions will greatly expand the technique's potential impact.

\bibliography{refs}
\bibliographystyle{iclr2020_conference}

\end{document}